\documentclass[final]{cvsports}

\usepackage{times}
\usepackage{epsfig}
\usepackage{graphicx}
\usepackage{amsmath}
\usepackage{amssymb}
\usepackage{multirow}
\usepackage{rotating}
\usepackage{comment}
\usepackage{textpos}
\usepackage{booktabs}

\usepackage[pagebackref=true,breaklinks=true,colorlinks,bookmarks=false]{hyperref}

\def\cvprPaperID{13} % *** Enter the CVPR Paper ID here
\def\confYear{CVsports 2021}

\begin{document}

%%%%%%%%% TITLE
\title{Camera Calibration and Player Localization in SoccerNet-v2 and \\ Investigation of their Representations for Action Spotting}

\author{
Anthony Cioppa*\\
{\small University of Li\`ege}\\
%{\tt\scriptsize anthony.cioppa@uliege.be}
\and
Adrien Deli\`ege*\\
{\small University of Li\`ege}\\
%Universit\'e de Li\`ege\\
%Liège, Brussels\\
%{\tt\scriptsize adrien.deliege@uliege.be}
\and
Floriane Magera*\\
{\footnotesize EVS Broadcast Equipment}\\
%Thuwal, KSA\\
%{\tt\scriptsize f.magera@evs.com}
\and
Silvio Giancola*\\
{\small KAUST}\\
%{\tt\scriptsize silvio.giancola@kaust.edu.sa}
%\phantom{{\tt\small adrien.deliege@uliege.be}}
\and
Olivier Barnich\\
{\small EVS Broadcast Equipment}\\
%\phantom{{\tt\small email}}
\and
Bernard Ghanem\\
{\small KAUST}\\
%\phantom{{\tt\small email}}
\and
Marc Van Droogenbroeck\\
{\small University of Li\`ege}\\
%Universit\'e de Li\`ege\\
%Liège, Brussels\\
%{\tt\small M.VanDroogenbroeck@uliege.be}
%Danmark\\
%{\tt\small mail@mail.dk}
%\and
%$^*$These authors contributed equally.
}

\maketitle

\newcommand{\B}{\bf}
\newcommand{\mysection}[1]{\vspace{2pt}\noindent\textbf{#1}}

\newcommand\blfootnote[1]{%
  \begingroup
  \renewcommand\thefootnote{}\footnote{#1}%
  \addtocounter{footnote}{-1}%
  \endgroup
}
\blfootnote{\textbf{(*)} Denotes equal contributions to this project. Contacts: anthony.cioppa@uliege.be, adrien.deliege@uliege.be, f.magera@evs.com, silvio.giancola@kaust.edu.sa. More at \url{https://soccer-net.org/}.}

%%%%%%%%% ABSTRACT
\vspace{-1cm}

\begin{abstract}
    % Catchup generic statement 
    Soccer broadcast video understanding has been drawing a lot of attention in recent years within data scientists and industrial companies. This is mainly due to the lucrative potential unlocked by effective deep learning techniques developed in the field of computer vision. 
    % What are we doing?
    In this work, we focus on the topic of camera calibration and on its current limitations for the scientific community. 
    %Motivation, insight
    More precisely, we tackle the absence of a large-scale calibration dataset and of a public calibration network trained on such a dataset. 
    %What we do exactly
    Specifically, we distill a powerful commercial calibration tool in a recent neural network architecture on the large-scale SoccerNet dataset, composed of untrimmed broadcast videos of 500 soccer games. We further release our distilled network, and leverage it to provide 3 ways of representing the calibration results along with player localization. Finally, we exploit those representations within the current best architecture for the action spotting task of SoccerNet-v2, and achieve new state-of-the-art performances.
    % Extra stuff
\end{abstract}

%%%%%%%%% BODY TEXT

%\begin{textblock*}{\textwidth}(0cm,-9.9cm) % {block width} (coords) %put x = 0.25 for left-align
%   \centering
%   {\footnotesize $^*$These authors contributed equally. Contacts: (anthony.cioppa, adrien.deliege)@uliege.be, f.magera@evs.com, silvio.giancola@kaust.edu.sa}
%\end{textblock*}

\section{Introduction}

\begin{figure}
    \centering
    \includegraphics[width=\linewidth]{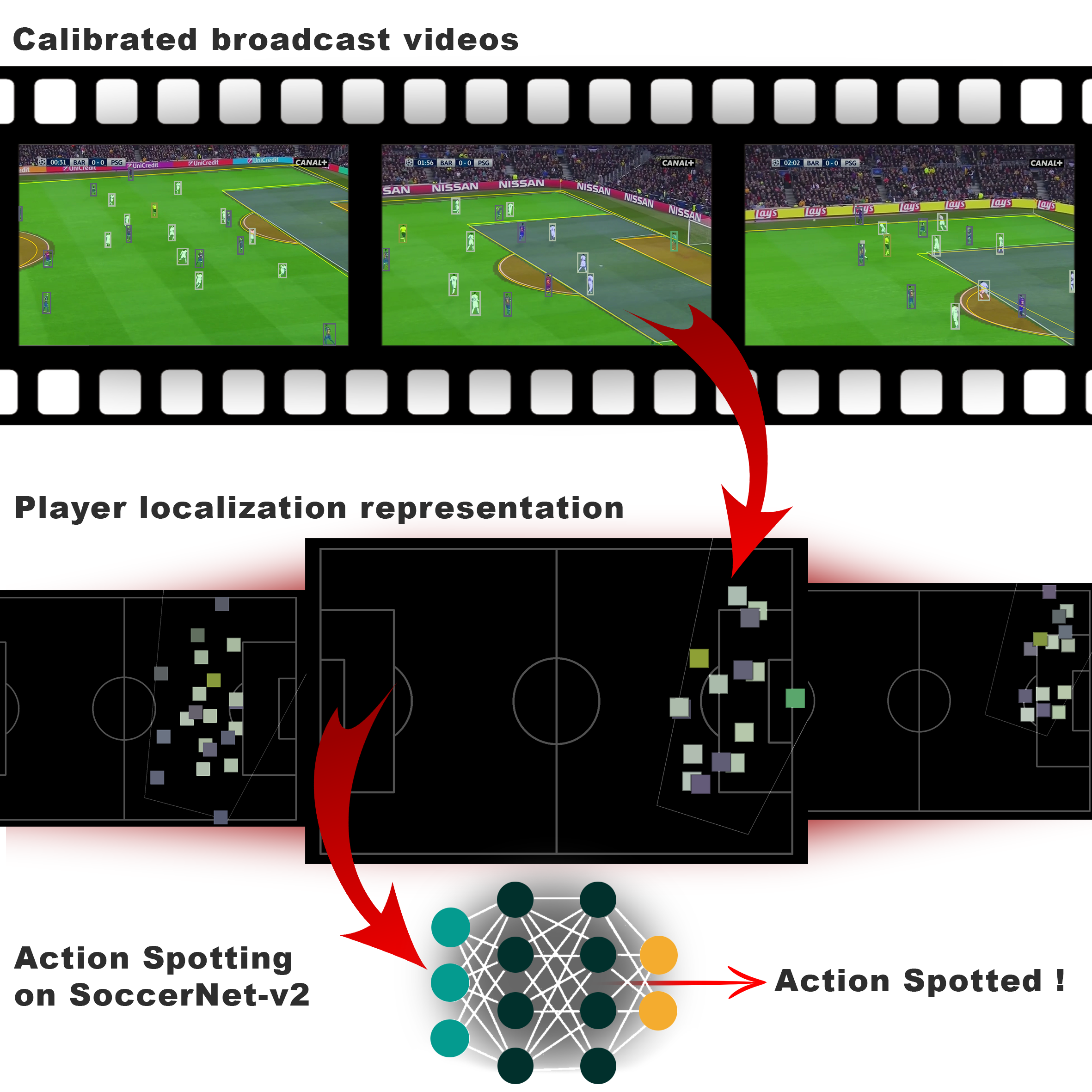}
    \caption{\textbf{Overview.} We compute and release the camera calibration parameters along player localization in real-world coordinates for the 500 soccer games of the SoccerNet dataset, we generate various types of calibration-based data representations, and we leverage them for the task of action spotting in SoccerNet-v2.}
    \label{fig:graphical_abstract}
\end{figure}

%General statement about soccer. 
Soccer is often regarded as one of the most unifying activities worldwide, with thousands of professionals entertaining millions of amateurs. Such a large audience makes soccer a very lucrative business, generating billions of dollars of revenue each year from broadcast events~\cite{statista}. The audiovisual data recorded during the games hides valuable insights about the players positions, the tactics, the strengths and weaknesses of each team. Hence, it is important for clubs and coaches to stay at the top of the data analytics wave, and for the fans, the data can be leveraged to provide customized services, such as personalized replays or enhanced player and game statistics. However, many general challenges of computer vision in sports have to be faced~\cite{moeslund2014computer,thomas2017computer}. Besides, the amount of data to process is so large that automated tools need to be developed. This explains the recent rise in deep learning algorithms to perform various tasks such as action spotting~\cite{cioppa2020context,Deliege2020SoccerNetv2,Giancola2018SoccerNet}, player counting~\cite{cioppa2020multimodal} and  tracking~\cite{Hurault2020Self,manafifard2017asurvey}, ball tracking~\cite{kamble2019adeep}, tactics analysis~\cite{Suzuki2019Team}, pass feasibility~\cite{Sangesa2020UsingPB}, talent scouting~\cite{decroos2019actions}, game phase analysis~\cite{Cioppa2018ABottom}, or highlights generation~\cite{agyeman2019Soccer,sanabria2019adeep}.

% Introducing calibration and its uses
In this work, we investigate the topic of camera calibration for researchers in computer vision focused on soccer. Camera calibration serves as a bridge between the images recorded and the physical world. It allows to project any point located on the field of the recorded frame to its real-world coordinates on a plane of the actual soccer field. It can thus provide knowledge about the part of the field recorded by the camera or the localization of the players on that field.  %Some of these tasks involve the knowledge of the part of the field displayed or the localization of the players on that field. Technically, this amounts to recovering the camera calibration, which allows to project a point on the field from the frame recorded to its location on a generic model of the field. 
One of the main commercial uses of camera calibration is the insertion of graphical elements in augmented reality. Inserting graphical elements may be used to ensure that the rules of the game are respected, such as automatic offside or goal line technologies \cite{EVSXeebra}. However, most common applications aim to improve the viewer experience by fancier storytelling and with game analytics \cite{VizLibero}. 

%Dig a hole
Given the value of camera calibration tools, it is not surprising that the best methods belong to private companies. This prevents scientific research on that topic to flourish at large scale. 
%Fill the hole
For that reason, %in agreement with EVS Broadcast Equipment, 
we leverage a powerful commercial tools~\cite{EVSXeebra} to train a neural network on the large-scale SoccerNet dataset~\cite{Giancola2018SoccerNet}, and we release the latter to the community, along with calibration estimates for the 500 complete games available. Furthermore, we propose 3 different ways of representing the player localization in real-world coordinates obtained from the camera calibration: a top view image of the game, a feature representation, and a player graph. From an application perspective, we investigate the use of calibration-related information for the task of action spotting in SoccerNet-v2~\cite{Deliege2020SoccerNetv2}. Those contributions are illustrated in Figure~\ref{fig:graphical_abstract} and further outlined below.

\mysection{Contributions.} We summarize our contributions as follows. 
\textbf{(i) Calibration for SoccerNet.} We provide calibration estimates and player localization for the 500 soccer games of the SoccerNet dataset, and we release the first calibration algorithm trained on such a large-scale soccer dataset.
\textbf{(ii) Data representations.} We provide top view image-based, compressed feature-based, and player graph-based representations of the calibration data and player localization.
\textbf{(iii) SOTA on action spotting in SoccerNet-v2.} As use case, we investigate the use of these representations in a state-of-the-art network for the action spotting task of SoccerNet-v2 and we improve its performances.

%Let's now focus on our problem. Dig a hole. Show there is a problem to be solved.
%Is it more challenging in football than in other sports? What are the specificities of camera calibration in football? In our case, we have broadcast videos, what are the extra challenges? 

%In a broadcast soccer video, retrieving the camera calibration is challenging because of the movement of the cameras, the use of multiple camera views, and the size of the field, which makes it impossible to capture it in a single view. 

%because soccer events occur in very dynamic environment, namely in terms of lighting and weather conditions. Automatic camera calibration for sports events usually relies on pitch detection. The sobriety/sparsity of the soccer's pitch markings makes it one of the most difficult environment for automatic single frame camera calibration (along with baseball).

%In this work, we focus on single frame camera calibration, as the temporal consistency between frames is often broken by camera change. 

%Fill the hole. Provide general trends/attempts to solve the problem, and why they do not solve it completely. Then, say what we do differently to avoid the problems of other methods, why we are better. 

%Precisely, what are our contributions? 3 contributions is good. Can be written a bit later when we have a better overview of what we put in the paper.
% \newpage
\section{Related work}

%More details on previous calibration methods and their problems, and what we do differently to avoid those problems. 

\mysection{Calibration.} In the context of sports events, camera calibration often benefits from the presence of a field whose layout is specified by the rules of the game. The camera may be parameterized using the full perspective projection model, but also using a homography model. Indeed, the field being most often planar, it is a convenient calibration rig to estimate the homography between the field plane and the image. %As broadcast cameras are often mounted on a tripod, the translation of the optical center of the camera is very small compared to the size of the scene. Thus, another suitable camera parameterization is the pan-tilt-zoom (PTZ) model, where both the camera position and roll are kept constant. 
Hereafter, ``camera calibration'' means the estimation of the intrinsic and extrinsic camera parameters.

For soccer, existing methods are assessed on the World Cup 2014 dataset \cite{Homayounfar2017Sports}, which introduces a metric based on the Intersection over Union (IoU) between the reference field model and its predicted deprojection from an image. This work %also proposes an original approach to camera calibration which 
leverages the segmentation of horizontal and vertical lines to derive a set of plausible field poses from the vanishing points, and selects the best field after a branch-and-bound optimization. However, it requires at least two of both vertical and horizontal lines to estimate the vanishing points. Some areas of the field contain few line markings, restricting the practical use of the method to goal areas. 

Another common approach is to rely on a dictionary of camera views. 
%Unlike approaches based on vanishing points, the scarcity of pitch markings in an image is not an issue here. Indeed, the search in the dictionary is pixel-based and not subject to any geometric conditions. Therefore, an image with, e.g., a single marking line passing through could make for a valid input image in this approach, although one should be wary of the ambiguities that such images would still carry. 
The dictionary associates an image projection of a synthetic reference field model to a homography used to produce said projection. Each input image is first transformed  %Upstream from the dictionary, this approach also requires some image processing to transform an input image 
to resemble a projection of the synthetic field, typically by a semantic segmentation of the field lines \cite{Chen2019Sports,Sharma2018Automated} or of the areas defined by the field lines \cite{Sha2020End}. That segmentation is then associated with its closest synthetic view in the dictionary, giving a rough estimate of the camera parameters, which is eventually refined to produce the final prediction. %The last step of this approach consists in refining the camera parameters starting from this first estimate.  
One drawback of this kind of approach is that the processing time scales poorly with the size of the dictionary. Some applications require a large dictionary, which may become a bottleneck if real-time processing is required. 

Some other calibration methods rely on tracking algorithms. Lu \etal~\cite{Lu2019PanTilt} use an extended Kalman filter to track the pan-tilt-zoom (PTZ) camera parameters. Citraro \etal~\cite{Citraro2020Real} use a particle filter to track the camera orientation and position. Due to the nature of tracking, these methods are restricted to deal with uncut, single-sequence video streams, making them inappropriate for a dataset of broadcast videos with many discontinuities, as in SoccerNet.
%Indeed, a broadcast video stream is composed of the video streams of multiple cameras covering the broadcast event, alternating the points of view every few seconds to give the viewer a better sense of the action. Therefore, broadcast video streams contain cuts with drastic changes of viewpoints every few seconds, and tracking-based methods are thus inadequate to be applied as is, since tracking would most likely be lost and in need of reinitiatilization at every cut. 

Kendall \etal~\cite{Kendall2015PoseNet} introduced the concept of training a neural network to directly predict the camera parameters from an image. This approach was further investigated successfully by Jiang \etal\cite{Jiang2020Optimizing} where the predicted homography is further refined by iterative differentiation through a second neural network that predicts the error. Due to the amount of computation needed in this latter step, this method is quite slow ($0.1$ fps). Sha \etal~\cite{Sha2020End} also use a neural network to refine the camera parameters found within the dictionary for the input image. They use a spatial transform network, trained to predict the homographic correction necessary to align two segmented images.  
In our work, we opt for the latter method because it does not involve tracking, reports a processing rate of up to 250 fps, and achieves good performances on the World Cup dataset.

%We now motivate our choice of a camera calibration method for our application of soccer events detection. As stated above, tracking-based methods are not appropriate  in our case. Moreover, we  should only consider methods able to process images at the native framerate of the cameras (i.e., in “real-time”). The method of Sha \etal \cite{Sha2020End} seems to be the best candidate as it (1) does not involve tracking, (2) reports processing rate of up to 250 FPS, and (3) gets good performance on the World Cup dataset. We therefore opt for this method in order to demonstrate the usefulness of camera calibration for our action spotting experiments.

%speak about datasets of calibration, and show there is only one small, and so our contribution is great, even if it is not perfect.

%[To insert somewhere in intro or related work] To enrich the dataset with even more actionable information, we release the calibration algorithm used in our experiments along with the pre-computed calibrations.

\mysection{Action Spotting.} The task of action spotting in soccer considered in this work was introduced by Giancola \etal~\cite{Giancola2018SoccerNet} along with the large-scale SoccerNet dataset. 
The objective is to identify at which moment various salient game actions occur, such as goals, corners, free-kicks, and more. Retrieving such information is valuable for downstream tasks such as camera selection in live game production, post-game soccer analytics, or automatic highlights generation. While detecting players on broadcast images can now be achieved with existing deep learning algorithms~\cite{Cioppa2019ARTHuS, He2017MaskR}, combining spatio-temporal information about their localization to infer the occurrence of game actions remains challenging as it requires a high level of cognition. Besides, in broadcast videos, several cameras are used and important actions are replayed, breaking the continuity of the stream. 

In SoccerNet~\cite{Giancola2018SoccerNet}, Giancola \etal focus on three types of actions: goals, cards, and substitutions, which are temporally annotated with single anchors to retrieve. Several baselines are proposed, all of which rely either on ResNet~\cite{He2016Deep}, I3D~\cite{carreira2017quo}, or C3D~\cite{tran2015learning} frame features computed at 2 frames per second followed by temporal pooling methods (NetVLAD and MaxPool), with the ResNet features yielding the best results. Several works followed, building on the same set of pre-computed ResNet features. Cioppa \etal~\cite{cioppa2020context} develop a particular loss function that takes into account the context surrounding the actions in the temporal domain. They use it to perform a temporal segmentation of the videos before using a spotting module, achieving state-of-the-art results. Similarly, Vats \etal~\cite{vats2020event} handle the temporal information around the actions with a multi-tower CNN that takes into account the noise due to the single anchor annotation scheme. Tomei \etal~\cite{tomei2020RMS} randomly mask a portion of the frames before the actions to force their network to focus on the following frames, as those may contain the most discriminative features to spot actions. By further fine-tuning the last block of the ResNet backbone, they achieve a strong state-of-the-art results on SoccerNet-v1. Rongved \etal~\cite{rongved-ism2020} directly learn a whole 3D ResNet applied to the video frames on 5-seconds clips. This turns out to be an ambitious approach with moderate results, given the huge volume of data to process from scratch. Vanderplaetse \etal~\cite{Vanderplaetse2020Improved} propose a multimodal approach by including audio features, first extracted with a pre-trained VGG-ish network, then averaged over 0.5 seconds windows and synced with the 2 fps original ResNet features. They are processed in parallel streams before undergoing a late fusion, yielding the best results in several action classes.

Besides those works, the literature is rich in papers using either small custom datasets, such as ~\cite{fakhar2019event,jiang2016automatic}, or focusing on event recognition from pre-cut clips and selected frames rather than spotting actions in untrimmed videos, such as~\cite{khan2018soccer,Khan2018Learning,khaustov2020recognizing}, or even targeting a single class, such as goals~\cite{Tsagkatakis2017GoalED}. In this work, we tackle the large-scale action spotting task of SoccerNet-v2, the extension of SoccerNet proposed by Deli{\`e}ge \etal~\cite{Deliege2020SoccerNetv2}. It covers 17 classes of actions, annotated for the 500 untrimmed SoccerNet games, and constitutes the most appropriate public benchmark for research on action spotting in soccer.

%\subsection{Action spotting based on calibration}

%Performing some kind of action spotting task with the help of a camera calibration algorithm is less common in the literature. Many calibration-focused papers do not aim to tackle downstream tasks, and research articles on action spotting do not have calibration algorithms readily available for their dataset. Still, 
%[To fill in]. Maybe say (and cite) that some works use player positions in top view and trajectories but obtained by direct measurements, or private datasets, rather than from a camera calibration from the videos themselves. That setting is quite rare I think, if not inexistant.
%Use of calibration to generate trajectories on a top view, then use for some action spotting in soccer (pass, reception, shot) on a private dataset : \cite{Sanford2020Group}, and in ice hockey NHL : \cite{Mehrasa2017Learning}

%\cite{Zhong2018Time} try to predict next action, given sequence of previous actions, in ice hockey.
% \newpage
\section{Calibration and player localization}

\begin{figure*}
    \centering
    \includegraphics[width=\linewidth]{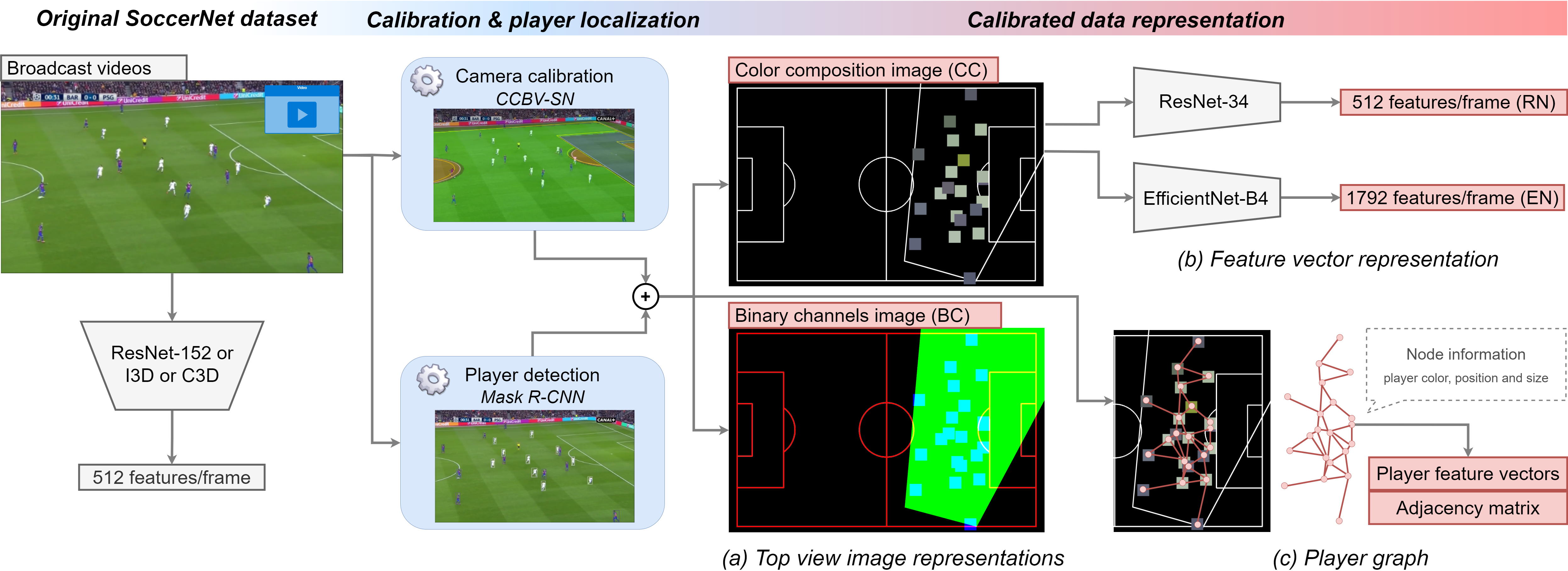}
    \caption{\textbf{Calibration and player localization representations.} The original SoccerNet dataset (left) provides raw videos of 500 complete soccer games as well as generic per-frame feature vectors. We distill a commercial calibration tool into a recent network architecture on SoccerNet, which we release along all the calibrations. We combine Mask R-CNN player detections with the calibration to provide 3 representations of the calibrated data, thus enriching the dataset with specific player-based information: (a) top view representations, (b) feature vectors representations, (c) a player graph representation. The red boxes, also released, further serve as inputs in neural networks to investigate the usefulness of calibration for the task of action spotting in SoccerNet-v2, leading to a new state-of-the-art performance.}
    \label{fig:main_figure}
\end{figure*}

\mysection{Contribution.} In SoccerNet~\cite{Giancola2018SoccerNet}, the frames of the raw videos are subsampled at 2 fps, then transformed into feature vectors, by passing through a ResNet-152~\cite{He2016Deep}, I3D~\cite{carreira2017quo}, or C3D~\cite{tran2015learning} network pre-trained on ImageNet~\cite{deng2009imagenet}, all of which are released with the dataset. Hence, those vectors only encode generic information about the frames. 
As first contribution, shown in Figure~\ref{fig:main_figure}, we enrich the SoccerNet dataset with actionable camera calibration estimates, along with players and referee localization. Such information provides a soccer-specific insight and is explicitly linked with the game in real-world coordinates. Besides releasing the largest set of calibration estimates to date, we are also the first to deliver a calibration algorithm trained on a large scale dataset such as SoccerNet. For synchronization purposes, we compute the calibration, player and referee localization for the 2-fps-subsampled set of frames considered in SoccerNet. In the following, we make no difference anymore between players and referees, all of which are called ``players'', and we call ``per-frame information'' any information computed for each of those subsampled frames.

\mysection{Calibration algorithm.} We base our calibration on the Camera Calibration for Broadcast Videos (CCBV) of Sha \etal \cite{Sha2020End}, but we write our own implementation, given the absence of usable public code. They use as calibration parameterization the homography between the field plane and the image, which is valid under the assumption of a planar field \cite{Hartley2004Multiple}. First, we describe their original algorithm, then we give the details of our changes. %Here, “camera calibration” means the estimation of the intrinsic and the extrinsic parameters of the camera, as described in the pinhole model. Sha \etal \cite{Sha2020End} use an equivalent parametrization: the homography between the field plane and the image is equivalent to the pinhole model parameterization under the assumption of a planar field \cite{Hartley2004Multiple}. 

The algorithm relies on a dictionary, \ie a set of pairs of artificial field zone segmentations, called ``templates'', and homographies. The dictionary is built in a pre-processing step, according to %the a priori knowledge of 
the camera parameters distribution over the training dataset. Since this distribution is unknown, it is estimated with a clustering algorithm based on Gaussian Mixture Models, that also determines the number of modes necessary to fit the distribution. The mean of each mode corresponds to a homography of the dictionary, that defines a camera perspective from which its corresponding template is generated as an artificial semantic image of the field.

%\mysection{Calibration algorithm.} %
CCBV itself consists of three steps, each performed by a specific neural network. 
First, a zone segmentation of the field is computed with a U-Net architecture~\cite{Ronneberger2015Unet}, where a zone is a field area enclosed by field lines. %The semantic zone segmentation is performed using a classical UNet architecture. 
Second, a rough estimate of the homography between the field plane and the image is obtained. A siamese network~\cite{Bromley1993Signature,Chicco2021Siamese} encodes the zone segmentation and the templates of the dictionary in feature vectors. The homography associated with the template encoding with the shortest $L^2$ distance to the zone segmentation encoding is the rough estimate of the sought homography. 
Third, this template homography is refined, in two steps. A Spatial Transform Network first regresses the homography between the zone segmentation and the template. %The STN is trained to ensure that the regression is valid both locally in the image and globally in top view. 
Then, the final homography prediction is obtained by multiplying the regressed homography with the template homography, giving the estimated calibration parameters.

%The loss function of the STN deserves some explanation. The objective of this loss is to ensure that the prediction is valid both locally in the image and globally in top view. It is a weighted sum of two terms. The first term is the dice loss between the ground truth segmentation mask and the warping of the reference model, using the refined homography prediction. The second term is the dice loss between the reference model and the warping in top view of ground truth segmentation masks, using the inverse of the refined  homography prediction. 

\mysection{Our training process.} Given the absence of a large-scale corpus of ground-truth camera calibrations in the literature, we opt for a student-teacher distillation approach. We consider a commercial tool~\cite{EVSXeebra} as teacher to generate a dataset of 12,000 pseudo-ground-truth calibrations on the SoccerNet dataset, which we use to train our student calibration algorithm.   %training and validation dataset. 
%We use thisWe train our student calibration algorithm as any regular neural network, but where the targets are provided by the teacher rather than a human annotator. 
Our training dataset is 60x larger than the World Cup 2014 dataset~\cite{Homayounfar2017Sports} used in~\cite{Sha2020End} and contains a larger variety of camera viewpoints, making our student calibration network a valuable candidate for universal camera calibration in the context of soccer. In fact, during the creation of the dictionary, more than 1000 modes are found by the clustering algorithm. Besides, during the training phase of the Spatial Transform Network, we notice vanishing gradient issues. To overcome this problem, we first pre-train it with a MSE loss and use leaky ReLU activations instead of ReLUs. After convergence, we compute the calibration estimates of the SoccerNet video frames with our trained calibration network. A binary score about the relevance of the calibration, set to 1 for frames with a plausible estimated calibration, is also computed by our student. This allows to discard cameras views that are not recorded by the main camera, such as close-up views, or public views. We release those estimates along our trained calibration network, which can be used with a wide variety of soccer videos. We denote CCBV-SN our student trained on SoccerNet.

\mysection{Player localization.} For each calibrated frame, we use Mask R-CNN~\cite{He2017MaskR} to obtain a bounding box and a segmentation mask per detected person instance. Then, we compute a field mask following~\cite{Cioppa2018ABottom} to filter out the bounding boxes that do not intersect the field, thus removing \eg staff and detections in the public. We use the homography computed by CCBV-SN to estimate the player localization on the field in real-world coordinates from the middle point of the bottom of their bounding box. Finally, we also store the average RGB color of each segmented player to keep track of a color-based information per person. As for the calibrations, we release this raw player-related information.

\section{Calibrated data representation}

\mysection{Contribution.} The calibration estimates and player localization are not easy to handle efficiently. Hence, to encourage their use in subsequent soccer-related works, we propose and release various easy-to-use representations of the calibration data extracted from the previous section. We illustrate these representations in Figure~\ref{fig:main_figure} and describe them in this section. We also discuss their pros and cons.

\subsection{Top view image representations}

In this section, we provide image representations of the player localization information. We use the calibration of CCBV-SN to generate a synthetic top view of the game containing generic field lines, the players represented by small squares, and the polygon delimiting the portion of the field seen by the camera. We represent that top view in two ways.
%In this section, we discuss how the player localization information is transformed into a usable representation. %We investigate two possibilities: a top view representation, and a graph-based one.

%\mysection{Top view.} As a first approach, we use the calibration to generate a synthetic top view of the game. It contains generic field lines, the players represented by small squares, and the polygon delimiting the portion of the field shown in the video. We test two ways of representing that top view.

\mysection{Color composition (CC).} We generate a RGB image where we first set field pixels in black and line pixels in white. Then, we superimpose with white pixels the contour of the polygon of the field seen by the camera. Finally, we represent the players by squares filled with their associated RGB color, overriding previous pixels in case of intersection. %In order to include a temporal dimension to the top view, we also represent, for each frame, the players of the two previous frames and of the two next frames, hence forming tracklets spanning 2 seconds of video.

\mysection{Binary channels (BC).} We generate an ``image'' composed of 3 binary channels: one for the generic field lines, one for the filled polygon of the field seen by the camera, and one for the players without their color information. 

%\mysection{Player graph.} Our second approach is to encode the player information in a graph, where nodes are players and vertices connect players close enough to each other, as follows.

%\textbf{(PG)} \textsl{Player graph.} ???

\mysection{Pros and cons.} A major advantage of image representations is their interpretability, since a human observer can directly understand relevant information about the game from such top views. Besides, they can be easily processed with convolutional neural networks in deep learning pipelines. As a drawback, they have a relatively large memory footprint compared with the low amount of actionable information that they actually contain. %This information scarcity may make them more prone to overfitting  when used in sophisticated neural networks.  
The color composition view has the advantage over the binary channels of keeping track of the color of the players, necessary for team discrimination and tactics analysis. On the other hand, the representation of a player in the binary channels is not influenced by a poor segmentation in the raw image or a color shift due to \eg an occlusion. Also, players located on field lines do not prevent those lines to be encoded properly in their binary channel, while they hide the lines in the color composition.

\subsection{Feature vector representation}

Inspired by Giancola \etal~\cite{Giancola2018SoccerNet}, we compress our top views as frame feature vectors extracted by pre-trained backbones. This is common practice in deep learning approaches, as universal networks trained on \eg ImageNet have an excellent transfer capability to encode meaningful visual information about any given image. 
We use top views of 224 $\times$ 224 pixels, with field lines of 4 pixels width 
and players of 8 $\times$ 8 pixels. We consider two backbones with similar number of parameters, both trained on ImageNet.

\mysection{ResNet-34 (RN).} This network has 21.8 million parameters and achieves 73.27\% top-1 accuracy on ImageNet. We use a frozen ResNet-34~\cite{He2016Deep} and collect the feature vectors of dimension 512 in its penultimate layer. 

\mysection{EfficientNet-B4 (EN).} This more recent network has 19 million parameters and achieves 82.6\% top-1 accuracy on ImageNet. We use EfficientNet-B4~\cite{tan2019EfficientNet}, which yields feature vectors of dimension 1792 in its penultimate layer.

\mysection{Pros and cons.} We choose these networks for their good trade-off between performance on ImageNet and inference time. Indeed, they allow for a much faster training of neural networks compared with the top views, as computationally expensive spatial convolutions have already been performed. As a drawback, the features collected from these networks are not interpretable anymore, which may reduce the possibilities of developing explainable models.

\subsection{Player graph representation}

\mysection{Player graph (PG).} Our third approach consists in encoding per-frame players information in a graph. Each player is represented with a node, whose features are defined by their associated RGB color, their position in real-world coordinates, and the area of the detected bounding box in the image frame. Two players are linked to each other with an edge if their real-world distance is below 25 meters, which we consider sufficient to pass contextual information between the nodes in the graph (\ie the players in the field).

\mysection{Pros and cons.} 
The player graph is a compromise between the compactness of feature representations and the interpretability of top views. Indeed, it explicitly encodes in a compact way the interpretable information that we want to embed in our descriptive features: the players color, their position in the field and their interactions with each other. 
Contrary to top view images, it does not encode any empty portion of the field, nor considers the field lines that are constant across the videos, which makes the learning focusing more on the interesting player features. 
The graph convolutional network (see next section) that processes the player graph aggregates features from neighboring players, which helps it understand real-world distances by discarding players further away. Yet, that aggregation does not consider different clusters of neighbors, which could lead to a misunderstanding between teammates and adversaries. 
\section{Experiments}

\mysection{Contribution.} In this section, we first validate with performance metrics the effectiveness of CCBV-SN as calibration algorithm. Then, we leverage our various calibration data representations in the particular use case of the action spotting task in SoccerNet-v2. We build on top of the current best network to achieve a new state-of-the-art performance.

\subsection{Validating the camera calibration distillation}

\begin{figure}
    \centering
    \includegraphics[width=\linewidth]{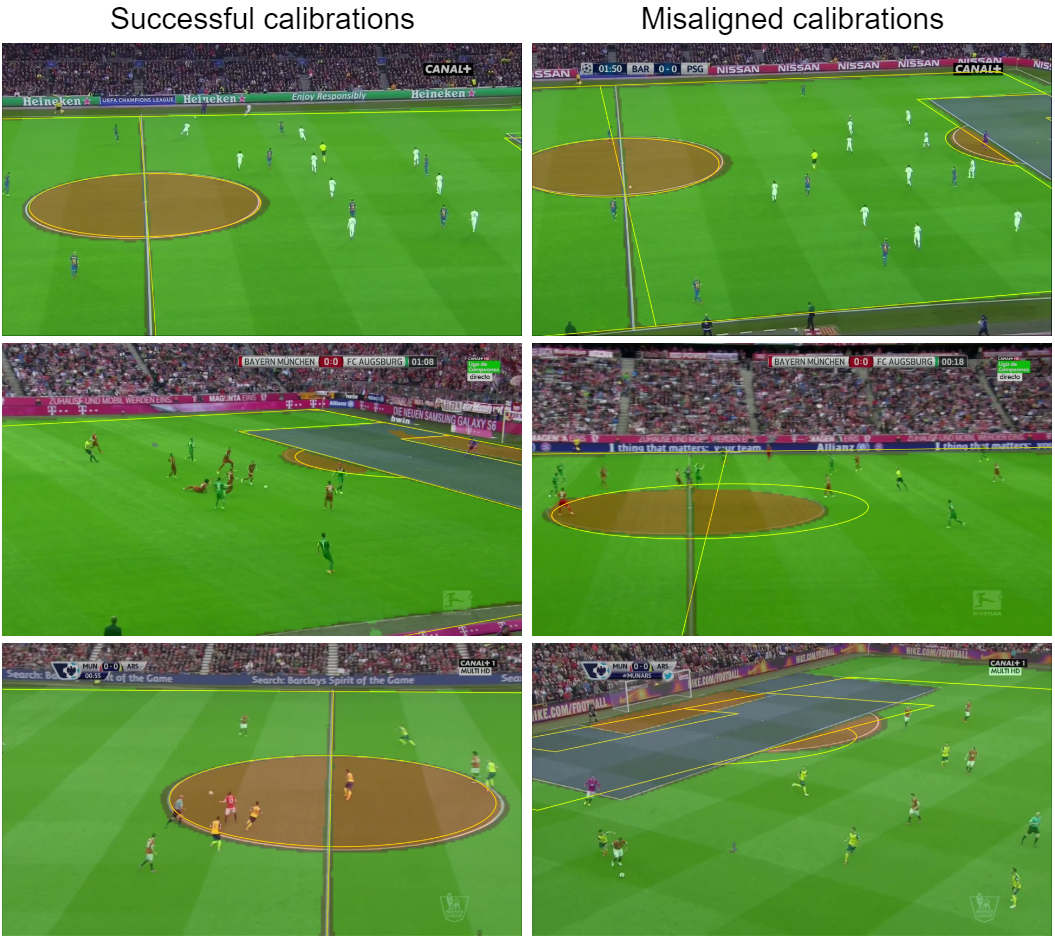}
    \caption{\textbf{Examples} of calibrations obtained with CCBV-SN. Globally, the results are satisfying and allow for an effective use of the calibration for downstream tasks such as action spotting.}
    \label{fig:calib_results}
\end{figure}

% Sentence to explain that we need to validate our student and, more importantly, our distillation pipeline (distill teacher on a dataset into student works well), to show that we can legitimately use it in our experiments. \textbf{\large Section under construction.}  Explain also difference between transfer learning (distill on soccernet, test on world cup) and distill on world cup.

In order to validate our calibration-based data representations and their use for an action spotting task, we first validate CCBV-SN as camera calibration algorithm.

%To validate the performances of both our teacher and student calibration techniques, we evaluate them on the World Cup 2014 dataset \cite{Homayounfar2017Sports}. None of our methods are trained on the World Cup 2014 dataset for the evaluation. Thus we actually measure the generalization capabilities of both methods on a completely new dataset. This evaluation also allows to quantify the performance drop induced by our distillation procedure.

\mysection{Dataset.} The World Cup 2014 dataset~\cite{Homayounfar2017Sports} stands as reference for evaluating soccer camera calibration methods. The test set comprises 186 calibrated images taken from 10 different games in various stadiums, perspectives, lighting conditions, and moments of the day. 

\mysection{Metric.} Following~\cite{Chen2019Sports, Sha2020End}, for each test image, we compute the entire intersection over union (\textit{IoU entire}) between the top view projections of the field model by the ground-truth camera and by the estimated camera, as well as the IoU restricted to the part of the field actually shown on the image (\textit{IoU part}). For both metrics, we report the mean and the median value across the test images. 

\mysection{Results.} We report the calibration performances in Table~\ref{tab:calibration-results}. The private teacher achieves the best results on 2 out of 4 metrics, which validates its use as a teacher in our distillation approach. It is topped by Citraro \etal~\cite{Citraro2020Real} on the IoU (entire) metrics, which finetune their method with additional manual annotations on the dataset. In comparison, none of our methods are trained on the that evaluation dataset. Thus we actually measure the generalization capabilities of our teacher and CCBV-SN on a completely new dataset. This evaluation also allows us to quantify the performance drop induced by our distillation procedure. CCBV-SN loses 6 to 12 points in the distillation process, making its performances close to~\cite{Sharma2018Automated}, especially on the IoU (part). This metric is actually the most relevant for us, as our use of the calibration is limited to the visible part of the field for the calibration data representations. Therefore, CCBV-SN is legitimately usable in the rest of our experiments, and is presumably even better on SoccerNet, since it is the dataset on which it has been trained. Some calibration results obtained with CCBV-SN are shown in Figure~\ref{fig:calib_results}.

\begin{table}[t]
\caption{\textbf{Calibration results} on the World Cup 2014 dataset~\cite{Homayounfar2017Sports}. The teacher tool outperforms the other methods on the IoU (part) metric. Our publicly released student network CCBV-SN, obtained by distilling the teacher in the CCBV architecture on SoccerNet, achieves acceptable transfer learning results.}
    \centering
    \vspace{0.3cm}
    \begin{tabular}{c|c|c|c|c}
          & \multicolumn{2}{|c|}{IoU (entire)} & \multicolumn{2}{c}{IoU (part)}  \\
         Method & Mean & Med. & Mean & Med. \\ \midrule 
         DSM~\cite{Homayounfar2017Sports} & 83 & - & - & - \\ 
         Sharma \etal~\cite{Sharma2018Automated} & - & - & 91.4 & 92.7 \\
         Chen \etal~\cite{Chen2019Sports} & 89.4 & 93.8 & 94.5 & 96.1 \\
         Sha \etal~\cite{Sha2020End}- CCBV & 88.3 & 92.1 & 93.2 & 96.1 \\
         Jiang \etal~\cite{Jiang2020Optimizing} & 89.8 & 92.9 & 95.1 & 96.7 \\
         Citraro \etal~\cite{Citraro2020Real} $^+$ & \B93.9 & \B95.5 & - & - \\
         Teacher~\cite{EVSXeebra} * & 91.7 & 93 & \B96.7 & \B98.7 \\
         Our CCBV-SN * & 79.8 & 81.7 & 88.5 & 92.3 \\ \midrule
    \multicolumn{5}{l}{\small{* no finetuning on WC14\ \  \ $^+$ used extra annotations on WC14}}
    \end{tabular}
    \label{tab:calibration-results}
\end{table}

\subsection{Use case: calibration-aware action spotting}

\begin{figure}
    \centering
    \includegraphics[width=\columnwidth]{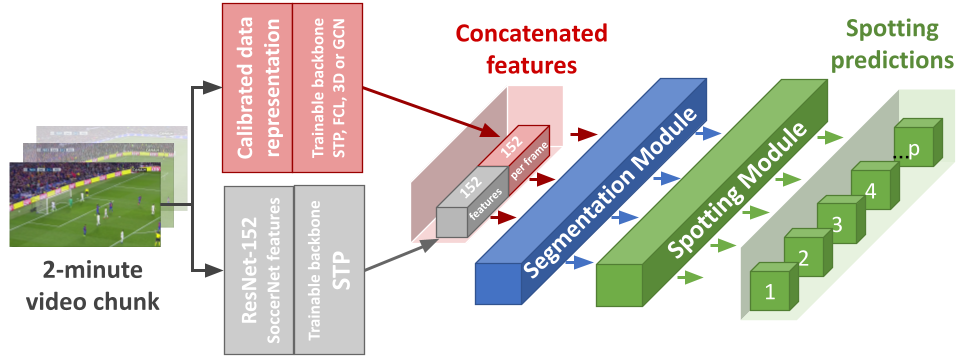}
    \caption{\textbf{Our action spotting pipeline} for the patterned actions. We include calibration information within CALF~\cite{cioppa2020context}, by concatenating frame feature vectors extracted from our various representations. This allows us to mix generic information from the SoccerNet features with player-specific information from the calibration. For each chunk, the network outputs $p$ spotting predictions.}
    \label{fig:method}
\end{figure}

In this section, we investigate a possible use case of our calibration representations, by leveraging a state-of-the-art network for the task of action spotting in SoccerNet-v2.

\mysection{Dataset.} The action spotting dataset of SoccerNet-v2~\cite{Deliege2020SoccerNetv2} consists in 110,458 action timestamps spread over 17 classes within the 500 complete games of the SoccerNet~\cite{Giancola2018SoccerNet} dataset, with 22,551 actions related to the 100 test games. Each action is annotated with a single timestamp, that must be retrieved as precisely as possible.

\mysection{CALF architecture.} We focus on integrating the calibration information along the original SoccerNet features in the Context-Aware Loss Function (CALF) architecture of Cioppa \etal in~\cite{cioppa2020context}. This architecture achieves state-of-the-art performances on the task of action spotting in SoccerNet-v2. As original features, we choose the ResNet features further reduced from 2048 to 512 components by PCA, as they yield the best results both in~\cite{cioppa2020context} and in ~\cite{Giancola2018SoccerNet}, which we also noticed in our preliminary experiments. 

CALF is composed of three trainable modules: a frame feature extractor, a temporal segmentation module, and an action spotting module. The first one is a convolutional spatio-temporal pyramid (STP) that aggregates the ResNet features across various time scales, and outputs a feature vector of user-defined dimension $d$ per frame. Our goal is to concatenate such features judiciously along frame feature vectors extracted from our calibration representations, as shown in Figure~\ref{fig:method}. The remaining two modules and the training protocol are kept as is to assess the improvement brought by only the calibration information.

\mysection{Processing our representations.} Each calibration data representation must be processed appropriately for a seamless integration within the network. We proceed as follows.

\textsl{Top views.} We process our top views with our own 3D-convolutional network \textbf{(3D)}. We choose the same structure as the STP module but where the kernels are extended to take into account the extra spatial dimension of the top view compared to the original ResNet features. The output is a $d$-dimensional vector for each frame that gathers the spatial and temporal information of the top view representation.   

\textsl{Feature vectors.} We investigate two ways of further processing the feature vectors obtained from the pre-trained backbones: (1) we use the trainable STP of CALF to extract $d$-dimensional frame feature vectors \textbf{(STP)}, (2) we fully connect our feature vectors through a trainable layer directly to feature vectors of dimension $d$ followed by a ReLU activation \textbf{(FCL)}. In the second case, we obtain per-frame feature vectors solely based on the raw frame information, without any temporal aggregation. 

\textsl{Player graph.} We design a graph convolutional neural network \textbf{(GCN)} to extract per-frame features from the player graph. For that purpose, we follow DeeperGCN~\cite{li2020deepergcn}. In particular, we build our architecture with 14 GCN blocks with residual skip connections. We leverage two layers of GENeralized Graph Convolution (GENConv) per block, that aggregate the lifted neighboring nodes using a softmax with a learnable temperature parameter. Then, a max operation across the node pools a global feature for the player graph. This feature is later lifted with a single fully connected layer to the desired dimension $d$.

\mysection{Class separation.} Intuitively, the player localization extracted with the calibration can prove more helpful for spotting some classes (\eg penalty) than others (\eg shot off target). Hence, we leverage our domain knowledge to split the 17 action classes of SoccerNet-v2 into two sets: ``patterned'' and ``fuzzy'' actions. We consider an action as ``patterned'' when its occurrence is systematically linked with typical player placements: penalty, kick-off, throw-in (one player outside the field), direct free-kick (player wall), corner, yellow card, red card, yellow then red card (players grouped around the referee for the card-related actions). On the other hand, a ``fuzzy'' action may occur in many different player configurations: goal, substitution, offside, shot on target, shot off target, clearance, ball out of play, foul, indirect free-kick. Given our class separation, we train two networks: one on the patterned classes that uses the calibration information and the original ResNet features, one on the fuzzy classes that only uses those ResNet features.

\mysection{Feature fusion.} For the network trained on the patterned classes, we input SoccerNet's ResNet features to the STM, collect $d$-dimensional feature vectors, and concatenate them with our $d$-dimensional vectors extracted by one of the above processing steps. This is illustrated in Figure~\ref{fig:method}. We set $d=$152, which allows us to simply plug a calibration-related branch next to the original branch of CALF working on SoccerNet's ResNet features. The concatenation yields feature vectors of dimension 304 and is performed just before the temporal segmentation module of the whole network. For the network trained on the fuzzy classes, we use SoccerNet's ResNet features only as in CALF, and set $d=304$ after the STM to have the same input dimension for the segmentation modules of the two networks.

\mysection{Training.} Following CALF, we process 2-minute video chunks. We extract frame feature vectors as described above, concatenate them when necessary, and input them to a temporal segmentation module, that provides per-class features and per-class actionness scores per frame. This module is trained with a context-aware loss that aggregates the temporal context around the actions. Those features and scores are concatenated and sent to an action spotting module, which provides predicted action vectors for the chunk, containing a localization estimate and a classification per predicted action. An iterative one-to-one matching connects those predictions with ground-truth action vectors, allowing to train the module with an element-wise MSE loss.

\mysection{Metric.} As defined in~\cite{Giancola2018SoccerNet}, we measure the action spotting performance with the Average-mAP. First, predicted action spots are said positive when they fall within a margin $\delta$ of a ground-truth timestamp from their predicted class. Then, the Average Precision (AP) is computed from Precision-Recall curves, then averaged over the classes (mAP). Finally, the Average-mAP is the AUC of the mAP obtained at margins $\delta$ varying from 5 to 60 seconds. Given our class separation, we merge the predictions of our two networks before computing the Average-mAP.
%Given our class separation and the linearity of this metric with respect to the classes, we can simply average across the classes the per-class Average-mAP obtained by our two networks. 
%Given our class separation and the linearity of this metric with respect to the classes, we report as final performance a weighted average of the performance of our networks, with 8/17 (resp. 9/17) of the Average-mAP achieved by the network trained on the patterned (resp. fuzzy) classes.

\begin{figure}
    \centering
    \includegraphics[width=\linewidth]{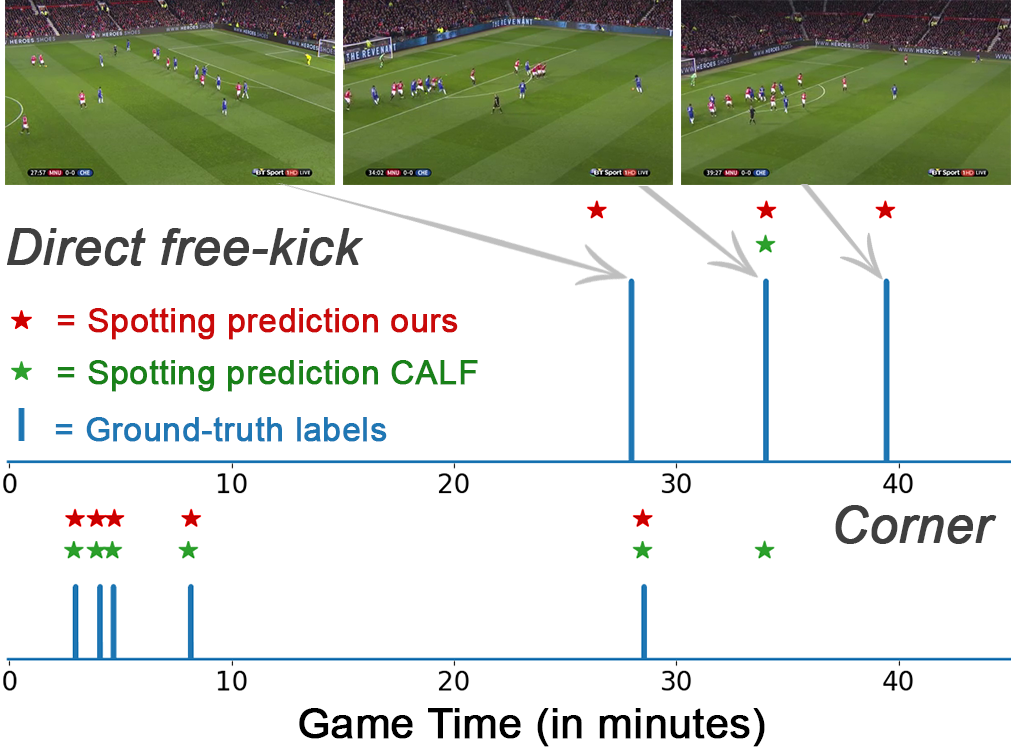}
    \caption{\textbf{Examples} of action spotting results on a game between Manchester United and Chelsea in December 2015. In this case, we spot correctly two more direct free-kicks than the original network, and we rightly avoid predicting a corner around 35 minutes.}
    \label{fig:quantitative_action_spotting}
\end{figure}

\mysection{Results.} We achieve our best result with the color composition reduced to frame features by ResNet-34 as calibration data representation, further bridged to $d$-dimensional feature vectors with a fully connected layer. This yields an Average-mAP of 46.8\% on the test set, reported in Table~\ref{tab:ActionSpotting-updated}, the current SoccerNet-v2 action spotting leaderboard. We achieve a novel state-of-the-art performance, outperforming the other methods by a comfortable margin. In particular, we prevail on 15 of the 17 classes, only topped by Vanderplaetse \etal~\cite{Vanderplaetse2020Improved} for kick-offs and penalties. Besides, kick-offs are the only actions for which our performances degrade compared to the original network, most probably because those actions are regularly unshown in soccer broadcasts~\cite{Deliege2020SoccerNetv2}. We illustrate some action spotting results in Figure~\ref{fig:quantitative_action_spotting}. We manage to spot actions that CALF misses, and some false positives of CALF are correctly avoided. On the current open competition of action spotting in SoccerNet-v2, organized on EvalAI, we achieve an Average-mAP of 46.4\% on the private challenge dataset. This validates the generalization capabilities of our network. 

\begin{table*}[ht]
\scriptsize
    \caption{\textbf{Leaderboard for action spotting} (Average-mAP \%) on SoccerNet-v2. Patterned actions are indicated with a * . We report the results of our best method, based on~\cite{cioppa2020context}, which outperform the other techniques on almost all the classes.}
    \centering
    \vspace{0.3cm}
    \setlength{\tabcolsep}{2.5pt}
    % \tabcolsep{6pt}
    \resizebox{\linewidth}{!}{
    \begin{tabular}{l||c||c|c|c|c|c|c|c|c|c|c|c|c|c|c|c|c|c}
    &  \begin{turn}{90}SN-v2\end{turn} & \begin{turn}{90} Ball out \end{turn} & \begin{turn}{90}Throw-in *\end{turn} & \begin{turn}{90}Foul \end{turn} & \begin{turn}{90}Ind. f.-kick \end{turn} & \begin{turn}{90}Clearance \end{turn} & \begin{turn}{90}Shot on tar. \end{turn} & \begin{turn}{90} Shot off tar. \end{turn} & \begin{turn}{90}Corner *\end{turn} & \begin{turn}{90}Substitution \end{turn} & \begin{turn}{90}Kick-off *\end{turn} & \begin{turn}{90}Yel. card *\end{turn} & \begin{turn}{90}Offside \end{turn} & \begin{turn}{90}Dir. f.-kick *\end{turn} & \begin{turn}{90}Goal \end{turn} & \begin{turn}{90}Penalty *\end{turn} & \begin{turn}{90}Yel.$\to$Red *\end{turn} & \begin{turn}{90}Red card *\end{turn} \\ 

       \midrule \midrule
Counts (test set)                      & 22551 & 6460&	3809&	2414&	2283&	1631&	1175&	1058&	999&	579&	514&	431&	416&	382&	337&	41&	14&	8 \\ \midrule \midrule
MaxPool~\cite{Giancola2018SoccerNet}     &  18.6 &   38.7 &  34.7 &  26.8 &  17.9 &  14.9 &  14.0 &  13.1 &  26.5 &  40.0 &  30.3 &  11.8 &   2.6 &  13.5 &  24.2 &   6.2 &  0.0 &0.9 \\ \midrule
NetVLAD~\cite{Giancola2018SoccerNet}    &  31.4 &  47.4 &  42.4 &  32.0 &  16.7 &  32.7 &  21.3 &  19.7 &  55.1 &  51.7 &  45.7 &  33.2 &  14.6 &  33.6 &  54.9 &  32.3 &  0.0 &  0.0 \\ \midrule
AudioVid~\cite{Vanderplaetse2020Improved}     &  39.9  &  54.3 &  50.0 &55.5 &  22.7 &  46.7 &  26.5 &  21.4 &  66.0 &54.0 &\B52.9 &  35.2 &  24.3 &46.7 &  69.7 &\B52.1 &  0.0 &  0.0 \\ \midrule
CALF~\cite{cioppa2020context}                  &40.7 &  63.9 &56.4 &  53.0 &41.5 &51.6 &26.6 &27.3 &71.8 &  47.3 &  37.2 &41.7 &25.7 &  43.5 &72.2 &  30.6 &0.7 &  0.7 \\ \midrule
Ours (CC + RN + FCL)                 &  \B46.8 &  \B68.7 &\B59.9 &  \B56.2 &\B45.5 &\B55.4 &\B32.5 &\B33.0 &\B78.7 &  \B60.4 &  34.8 &\B50.4 &\B33.6 &  \B48.6 &\B76.2 &  50.5 &\B3.1 &  \B8.5 \\ \bottomrule

%\multicolumn{22}{l}{Other SoccerNet-v1 results but with no public code available: Rongved \etal~\cite{rongved-ism2020}: 32.0 ; Vats \etal~\cite{vats2020event}: 60.1 ; Tomei \etal~\cite{tomei2020RMS}: \textbf{75.1}.}

    \end{tabular}}
    \label{tab:ActionSpotting-updated}
\end{table*}

For completeness, we give additional results with the different combinations of calibration data representation and feature extraction in Table~\ref{tab:ablations}. We see that the color composition with the 3D network and the player graph representation yield performances that are practically equivalent to our best result, while other variants are less effective. Hence, each calibration data representation is able to reach competitive performances. We do not report any result with extracted feature representations from top view images composed of binary channels as they globally yield much lower performances. Finally, fusing features from the top view and the player graph does not appear useful either as these contain essentially the same type of information.

\begin{table}[t]
\caption{\textbf{Action spotting results} (Average-mAP \%) obtained with our various data representations, feature vectors, and networks.}
    \centering
    \vspace{0.3cm}
    \begin{tabular}{c|c|c|c}
    Data repres. & Features & Network & Av.-mAP \\ \midrule
    Binary channels & - & 3D & 44.7\\
    Color compos. & - & 3D & 46.7\\
    Color compos. & ResNet-34 & STP & 43.5\\
    Color compos. & ResNet-34 & FCL & \B46.8\\
    Color compos. & Effic.Net-B4 & STP & 42.5\\
    Color compos. & Effic.Net-B4 & FCL & 45.5\\
    Player graph & - & GCN & 46.7\\ 
    \end{tabular}
    \label{tab:ablations}
\end{table}

%Somehow we'll need to say how we handle calibration for frames like close-up view... A reviewer or reader might ask the question.

% \newpage
%\input{sections/5_discussion}
% \newpage
\section{Conclusion}

In this paper, we examine the problem of computing, representing, and exploiting the camera calibration information for the large-scale SoccerNet dataset, composed of 500 soccer games. We leverage a powerful commercial tool to generate pseudo ground truths and manage to distill it into a recent deep learning algorithm. As first contribution, we release our distilled network, which is the first public soccer calibration algorithm trained on such a large dataset, along with its calibration estimates for the SoccerNet videos to enrich the dataset. We use our calibration and a player detection algorithm to obtain the player localization in real-world coordinates. To further serve the scientific community, our second contribution is to provide three actionable ways of representing those calibration data: top view images, feature vectors representations, and player graphs. Eventually, we investigate the benefit of using these representations in a deep learning network for the task of action spotting in SoccerNet-v2. Standing for our third contribution, we design an appropriate concatenation of generic video and specific calibration information within the current best network to achieve a novel state-of-the-art performance.

\mysection{Acknowledgments.} This work is supported by the DeepSport project of the Walloon Region and the FRIA, EVS Broadcast Equipment, and KAUST Office of Sponsored Research (OSR) under Award No. OSR-CRG2017-3405.
\clearpage

{\small
\bibliographystyle{ieee_fullname}
%\bibliography{bibliography}
\typeout{}\bibliography{bibliography}
}
%Use the following line if bibliography is not parsed
%\typeout{}\bibliography{bibliography}

\end{document}